\documentclass[letterpaper, 10 pt, conference]{ieeeconf}  

\IEEEoverridecommandlockouts                              

\usepackage{graphics} 
\usepackage{epsfig} 
\usepackage{times} 
\usepackage{amsmath} 
\usepackage{amssymb}  

\usepackage{amsfonts}
\usepackage{multirow}
\usepackage{booktabs}
\usepackage{graphicx}
\usepackage{subfigure}
\usepackage{mwe}
\usepackage{threeparttable}
\usepackage{CJKutf8}
\usepackage{tikz}
\usetikzlibrary{trees}
\usepackage{cite}

\usepackage{hyperref}
\hypersetup{
    colorlinks=true,
    linkcolor=blue, 
    citecolor=blue, 
    urlcolor=blue 
}

\usepackage{color}

\title{\LARGE \bf 
SLABIM: A SLAM-BIM Coupled Dataset in HKUST Main Building
}

\author{Haoming Huang, Zhijian Qiao, Zehuan Yu, Chuhao Liu, Shaojie Shen, Fumin Zhang and Huan Yin
\thanks{This work was supported in part by the HKUST-DJI Joint Innovation Laboratory, in part by the HKUST-BDR Joint Research Institute (Grant No. OKT25EG04), and in part by the Hong Kong Center for Construction Robotics (InnoHK center supported by Hong Kong ITC).}
\thanks{Haoming Huang is with the Division of Emerging Interdisciplinary Areas, Hong Kong University of Science and Technology, Hong Kong SAR.}
\thanks{Zhijian Qiao, Zehuan Yu, Chuhao Liu, Shaojie Shen, Fumin Zhang and Huan Yin are with the Department of Electronic and Computer Engineering, Hong Kong University of Science and Technology, Hong Kong SAR. }
\thanks{Corresponding author: Huan Yin}
}

\begin{document}

\tikzstyle{every node}=[draw=black,thick,anchor=west]
\tikzstyle{selected}=[draw=red,fill=red!30]
\tikzstyle{optional}=[dashed,fill=gray!50]

\maketitle
\thispagestyle{empty}
\pagestyle{empty}

\begin{abstract} 
Existing indoor SLAM datasets primarily focus on robot sensing, often lacking building architectures. To address this gap, we design and construct the \textit{first} dataset to couple the SLAM and BIM, named SLABIM. This dataset provides BIM and SLAM-oriented sensor data, both modeling a university building at HKUST. The as-designed BIM is decomposed and converted for ease of use. We employ a multi-sensor suite for multi-session data collection and mapping to obtain the as-built model. All the related data are timestamped and organized, enabling users to deploy and test effectively. Furthermore, we deploy advanced methods and report the experimental results on three tasks: registration, localization and semantic mapping, demonstrating the effectiveness and practicality of SLABIM. We make our dataset open-source at \url{https://github.com/HKUST-Aerial-Robotics/SLABIM}.
\end{abstract}

\section{Introduction} 
\label{sec:introduction}



 
The mobile robotics community has made significant efforts in building maps for autonomous navigation, with simultaneous localization and mapping (SLAM)~\cite{dissanayake2001solution,cadena2016past} being the most renowned technique. Over the past two decades, SLAM has achieved notable success, evolving from single sensor-based methods to sensor fusion schemes \cite{xu2022fast}, and advancing from purely geometric to high-quality metric-semantic mapping~\cite{rosinol2020kimera}. Intuitively, human beings can navigate using prior building architectures, such as 2D floor plans or semantic-enhanced 3D building information modeling (BIM). Recent studies have proposed achieving robotic navigation on such architectures~\cite{yin2023semantic,chen2024f3loc}, thereby reducing the need for high-quality maps from SLAM.

On the other hand, SLAM is gaining popularity in the architectural, engineering, and construction (AEC) industry~\cite{helmberger2022hilti} with the demand for high-quality modeling. Traditional stationary laser scanners often incur high costs and offer low efficiency in providing dense point clouds. Conversely, advanced SLAM systems deliver high-quality and colored dense modeling by mobile mapping for practical uses, such as quality assessment~\cite{kim2015framework} and digital twin generation~\cite{wang2024omni}.

\begin{figure}[t]
    \centering
    
    \includegraphics[width=0.95\linewidth]{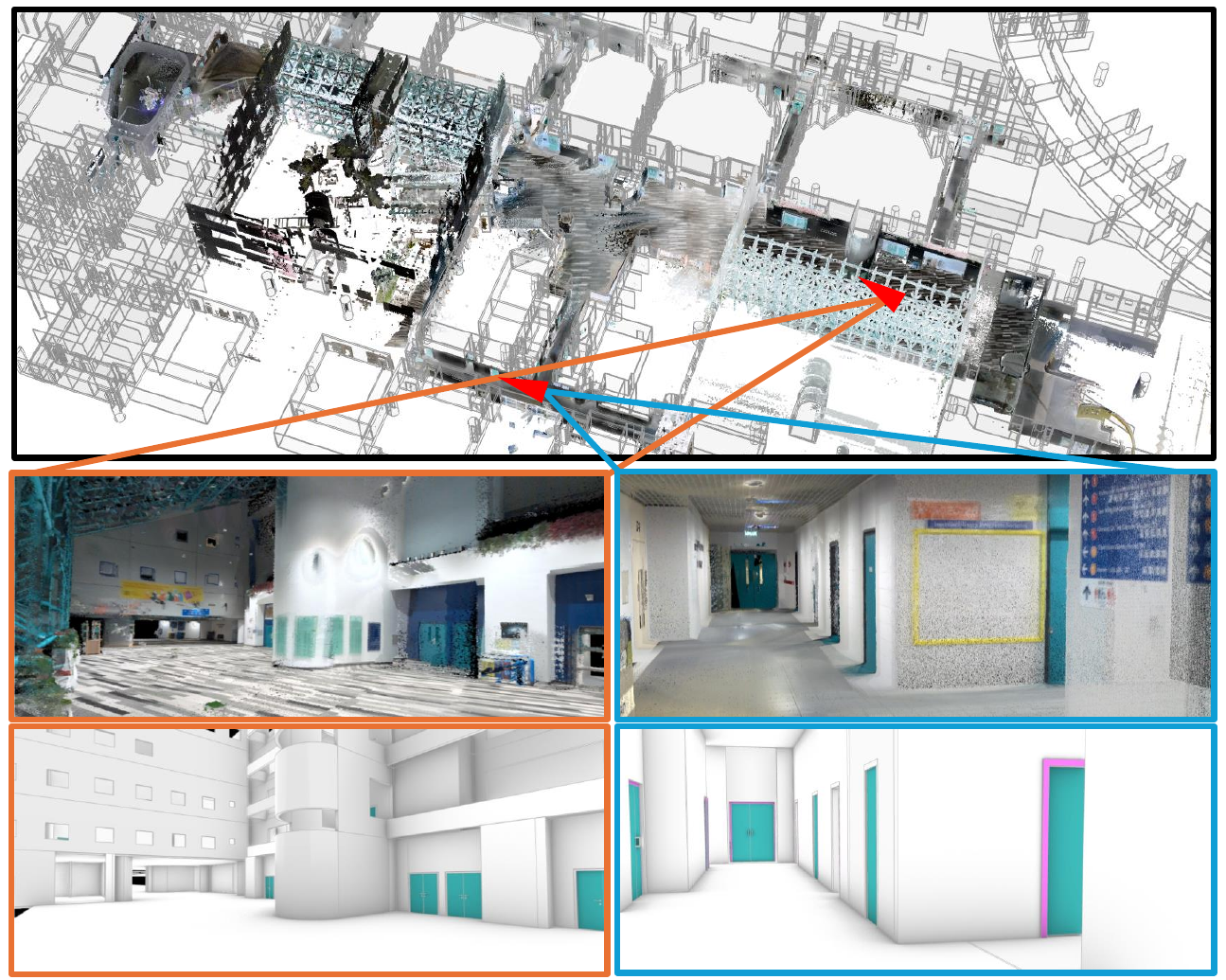}
    \caption{Screenshot of SLABIM in the HKUST main building. The top image shows a SLAM-generated colored map aligned with the BIM. The bottom images in boxes show the same views of real-world colored map and rendered images generated by SLAM and BIM, respectively. }
    
    \label{fig:teaser_image}
\end{figure}

We can observe a growing \textit{connection} between building architectures and SLAM, as the maps they generate represent as-designed and as-built models of the environments, respectively. This shared orientation and similar essence promotes promising directions in both communities. However, there is a lack of public datasets encompassing SLAM-oriented robotic data and corresponding building architectures. In this paper, we introduce a new dataset, SLABIM, to couple SLAM and BIM. Specifically, the BIM of SLABIM models a university building located on the Hong Kong University of Science and Technology (HKUST) campus. The SLAM-oriented data is a multi-session multi-sensor dataset and SLAM-generated maps in the same building. Figure~\ref{fig:teaser_image} presents a view of the two modalities. Additionally, we employ three practical tasks on SLABIM: global LiDAR-to-BIM registration, robot pose tracking on BIM and semantic mapping evaluation. We hope the SLABIM dataset will facilitate interdisciplinary research across various fields. In summary, the key contributions of this paper are twofold:
\begin{itemize}
    \item To our best knowledge, SLABIM is the \textit{first} open-sourced dataset that couples SLAM and BIM.
    \item We validate SLABIM with three different tasks, reporting experimental results and our findings.
\end{itemize}




\section{Related Work} 
\label{sec:related work}

\textbf{BIM Meets Robotics.} Early works by Boniardi \textit{et al.}~\cite{boniardi2016autonomous,boniardi2019pose} employed drawing maps and floor plans to facilitate robot navigation. LiDAR and visual-based robot localization (pose tracking) techniques were introduced in \cite{li2020online} and \cite{chen2024f3loc}, respectively. FloorplanNet by Feng \textit{et al.}~\cite{feng2023floorplannet} proposed learning features for global matching, thus enabling cross-modality global localization. Unlike traditional computer-aided design
(CAD) or floor plans, BIM encapsulates multi-dimensional information beyond mere geometrics. Studies by Yin \textit{et al.}~\cite{yin2022towards,yin2023semantic} transformed BIM into dense and semantic point clouds to facilitate LiDAR pose tracking. Hendrikx \textit{et al.}~\cite{hendrikx2021connecting} utilized feature points in BIM for robot localization. In addition to mobile robot localization, path planning was also achievable on BIM-based maps~\cite{hamieh2020bim,kim2022bim}.

Robotic techniques also benefit the BIM-aided construction automation. A promising direction is aligning SLAM-generated maps with BIM for construction purposes, focusing on aligning the two different modalities. The study in~\cite{lu2021novel} proposed extracting paths to achieve alignment for visual inspection on BIM. Our recent work~\cite{qiao2024speak} designed triangular descriptors for frame alignment between LiDAR maps and BIM. Another popular topic is converting as-built structures to BIM, known as Scan2BIM, for digital twin generation~\cite{wang2022object,wang2024omni}. The proposed SLABIM dataset provides both sensor data and BIM, allowing for the verification of most methods mentioned above.

\textbf{Indoor Datasets.} Indoor environments are common scenarios for mobile robots. Early datasets provided 2D laser scans to advance SLAM research~\cite{Radish}. RGB-D and camera-based sensing are also popular choices for indoor SLAM. Classical visual-aided datasets, such as 7-Scenes~\cite{shotton2013scene} and the TUM RGB-D dataset~\cite{sturm2012benchmark}, offered tracked frames and ground truth poses for evaluation. With the development of sensor technologies, various sensors have been designed to enhance robot perception and navigation. Consequently, recent indoor datasets focus on multi-sensor fusion and advanced tasks, such as FusionPortable~\cite{jiao2022fusionportable}, HILTI SLAM Challenge~\cite{helmberger2022hilti}, and THUD~\cite{tang2024mobile}. Notably, few datasets provide building architectures, though they can serve as typical scene priors for mobile robots. CubiCasa5K by Kalervo \textit{et al.}~\cite{kalervo2019cubicasa5k} provided large-scale floor plan images for building architecture parsing but lacks real-world sensor data. In this paper, we introduce SLABIM, which integrates both SLAM-oriented sensor data and building architectures.

\section{The SLABIM Dataset} 
\label{sec:ovewview}
This section first presents the HKUST BIM, which encodes the main building (Section~\ref{sec:BIM}). We then introduce the data acquisition platform used for collecting robotic sensing data (Section~\ref{sec:platform}), and detail the multi-session data collected in the main building (Section~\ref{sec:multi-session}).

\begin{figure}[t]
    \centering
    \subfigure[3D HKUST BIM and elements]{\includegraphics[width=1\linewidth]{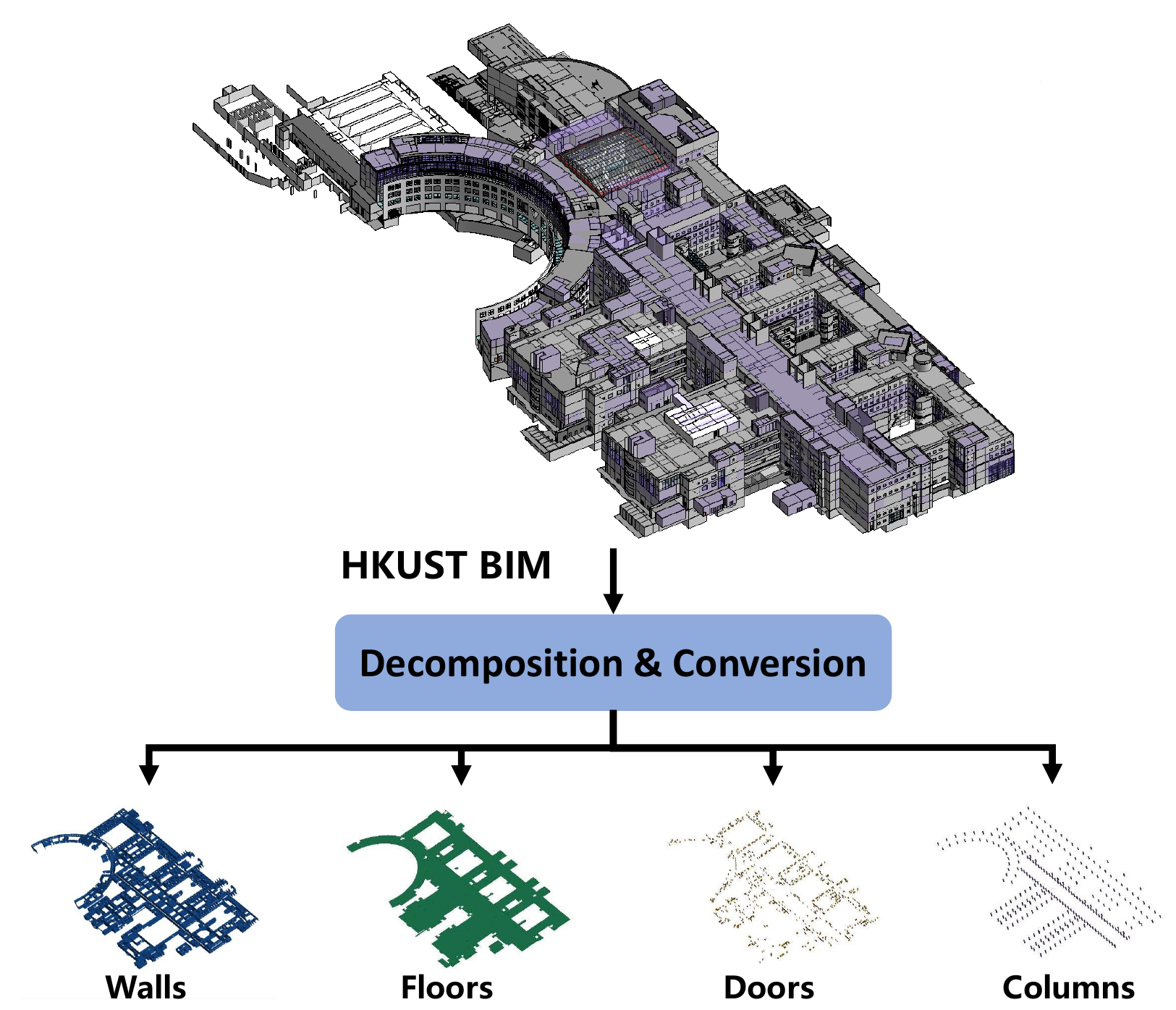}
    \label{fig:process}
    }
    \subfigure[2D floor plan of the 3rd floor (3F)]{\includegraphics[width=1\linewidth]{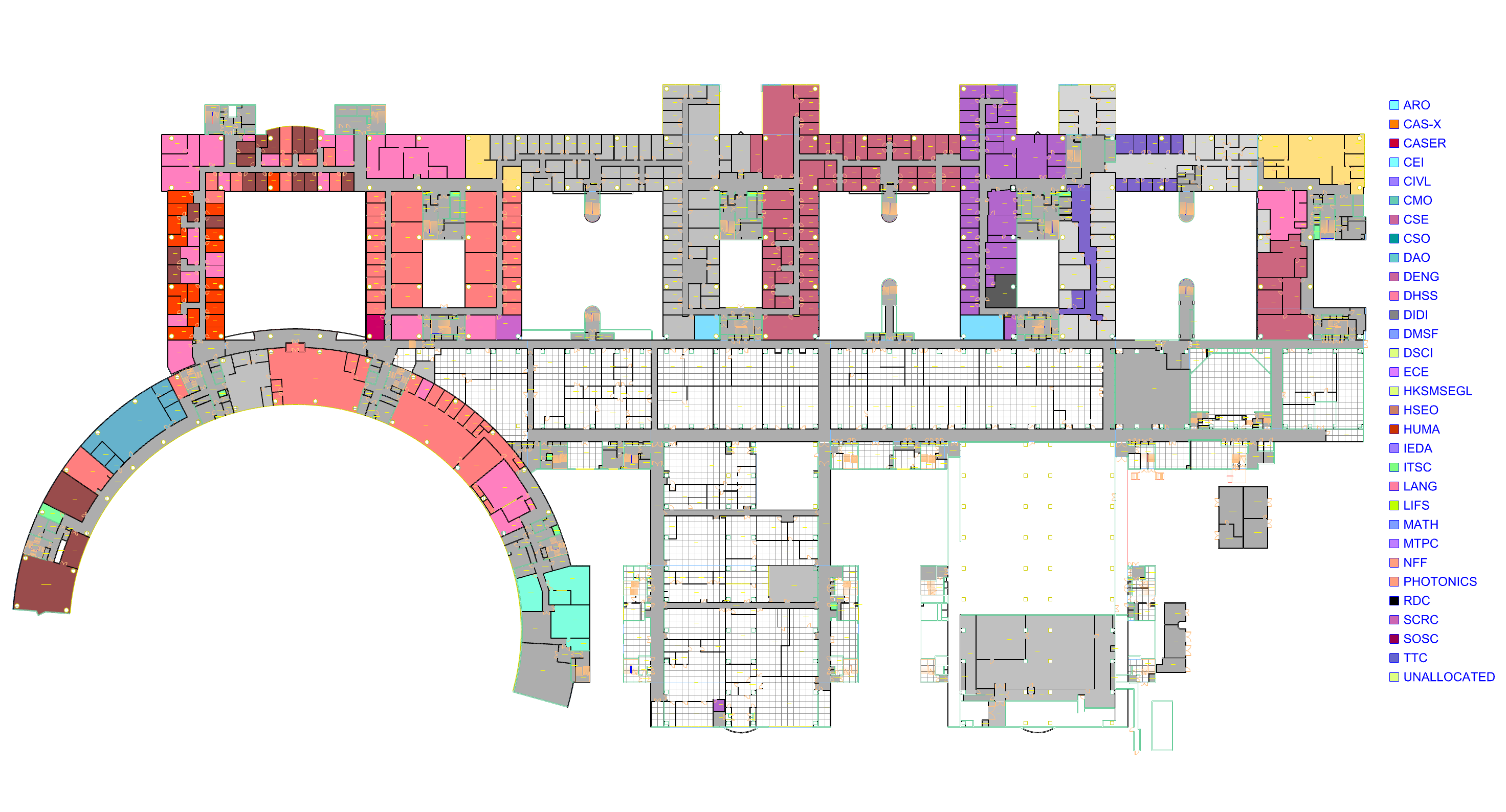}
    \label{fig:floorplan}
    }
    \caption{(a) HKUST BIM and the four typical types of elements extracted from the decomposition and conversion: walls, floors, doors, and columns. (b) 2D floor plan map exported from HKUST BIM, which could be used for map-based navigation. The column on the right shows the different departments (regions) of the university, e.g. ECE (Electronic and Computer Engineering).}
    \label{fig:HKUST_BIM}
\end{figure}

\subsection{HKUST BIM}
\label{sec:BIM}

The BIM model of this dataset is a part of the digital twin project in HKUST~\cite{chen2022building}, named HKUST BIM in the rest of this paper. The HKUST BIM covers the HKUST main building from LG7 (Lower Ground 7) to 7F (Floor) separately. The total floor area exceeds $2.2\times10^5$ m\textsuperscript{2}, featuring various types of offices, classrooms, gardens, lounges, and corridors. 

We refine the original HKUST BIM model to align with the practical requirements. This process involves two primary steps: decomposition and conversion. Firstly, the complete BIM is partitioned into individual BIMs for each floor. These decomposition operations are performed using Revit, a widely recognized professional software in the AEC industry. Additionally, we extract typical elements from the HKUST BIM, including walls, floors, doors, and columns, which can be regarded as semantic objects for advanced SLAM systems~\cite{rosinol2020kimera,liu2024slideslam}. Notably, the decomposition is automated through scripts in Revit, utilizing storey information and semantic tags in BIM. The decomposed BIM does not provide SLAM-friendly data for robotics. Therefore, the second step focuses on converting the elements and BIM into geometric data. We offer the converted mesh files in the public dataset due to the applicability and flexibility of this representation. 

In addition to the mesh files, we provide the original BIM and exported 2D floor plans, as shown in Figure~\ref{fig:HKUST_BIM}. The floor plans are stored in CAD format with annotations. Users can also convert the mesh files into other map representations, such as point clouds or segmented lines, thereby extending the applicability of the HKUST BIM. In Section~\ref{sec:applications}, we present applications using converted point clouds and report the experimental results.

\subsection{Dataset Acquisition Platform}
\label{sec:platform}



\begin{figure}[t]
    \centering
    \includegraphics[width=1\linewidth]{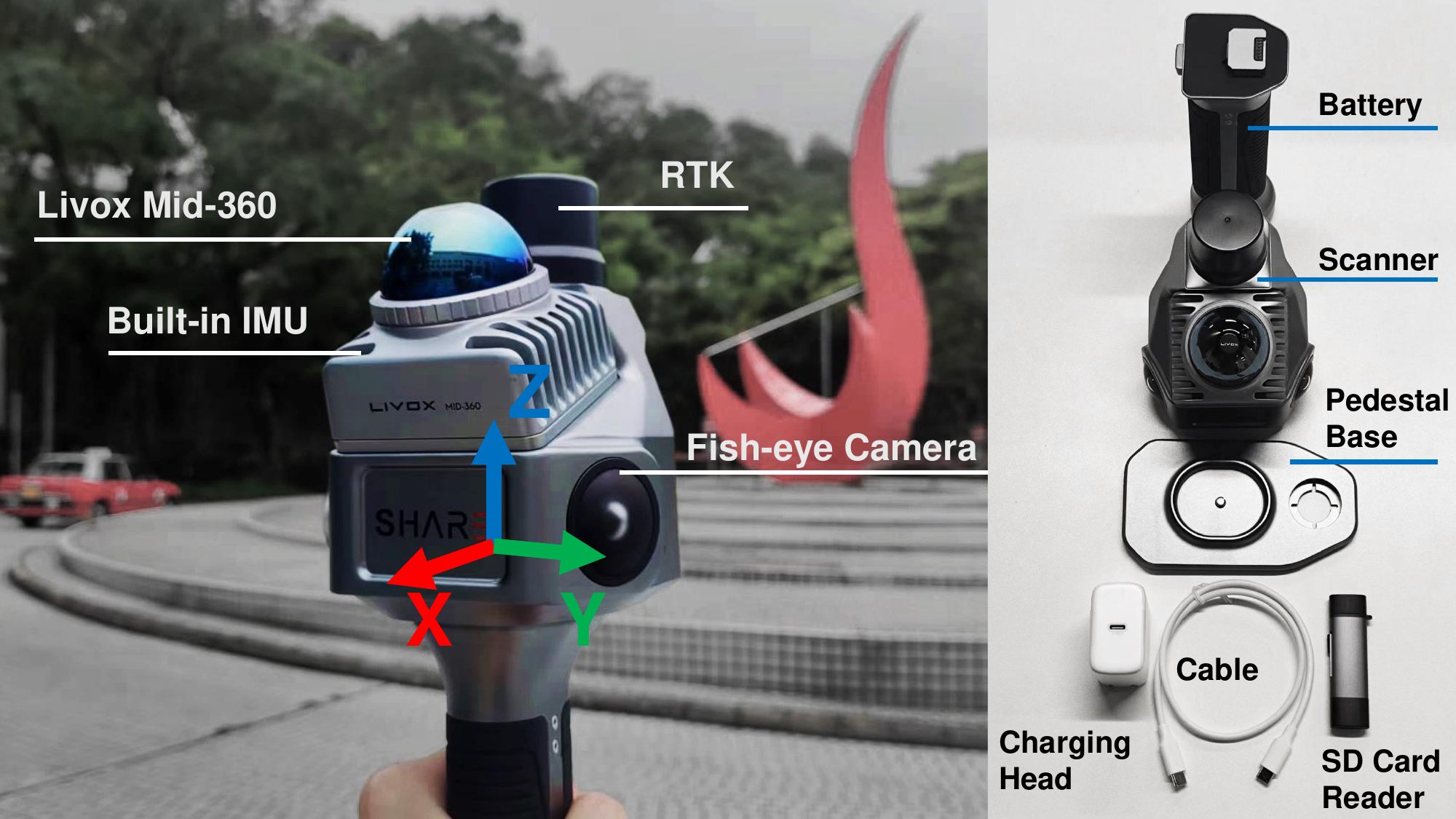}
    \caption{Sensor suite for data collection. A real-time SLAM system is integrated to produce colored 3D point cloud maps, with offline map refinement available to enhance mapping quality. }
    \label{fig:platform}
\end{figure}

The other part of SLABIM is the multi-session data collected by a handheld sensor suite. Figure~\ref{fig:platform} displays the sensor suite for data acquisition. The sensor configurations are described as follows:
\begin{enumerate}
    \item Fish-eye cameras. Two high-resolution fish-eye cameras are mounted on either side of the platform, arranged at 135° regarding the $x$-axis to ensure a wide field of view.
    \item 3D LiDAR scanner. A Livox Mid-360 LiDAR sensor is integrated to deliver precise measurements of the surrounding environment. The LiDAR is tilted 20° toward the ground, enabling comprehensive coverage of the ground, forward areas, and ceilings in the building.
    \item Inertial measurement unit (IMU). We use the built-in IMU in Livox directly.
    \item Real-time kinematic positioning (RTK). The RTK module provides precise geographic coordinates for indoor scanning in GPS-denied environments. When returning to areas with stable RTK signals, it automatically corrects system errors, thus improving the mapping accuracy.
\end{enumerate}

A more detailed summary of the characteristics can be found in Table \ref{tab:sensors}. With the sensors above, an internal sensor fusion-based SLAM system can provide real-time poses and colored point cloud maps in real time. The map colorization is based on the well-calibrated sensors. After collecting each session, an offline mapping system can also provide fine-tuned maps by specialized map refinement modules, including pose graph optimization and bundle adjustment. 

 

\begin{table}[!t]
\centering
\caption{Sensor Suite and Characteristics}
\label{sensor characteristics}
\renewcommand\arraystretch{1.2}
\resizebox{\columnwidth}{!}{%
\begin{tabular}{l|l|l}
\hline
\hline
\multirow{9}{*}{Sensors} & Basics & Size: 297.7$\times$103.7$\times$104.8mm; \\
 &  & Weight: 1010g; \\
 \cline{2-3} 
 & Camera & Resolution: 3040$\times$4032; \\
 &  & Field of View: 180$^\circ$ vert., 135$^\circ$ horiz.; \\
 & LiDAR & Density: ~$2\times10^5$ points/s; \\
 &  & Frequency: 10Hz; \\
 & IMU & ICM40609; Frequency: 200Hz; \\
 & RTK & Unicore UM980; Frequency: 10Hz; \\
 \hline
\multirow{4}{*}{Output} & Maps & Uncolorized: pcd format; Colorized: las format; \\
 & Odometry & Poses with timestamps in csv format \\
 & Rosbag & Raw data contained \\ 
 \hline
 \hline
\end{tabular}%
\label{tab:sensors}
}
\end{table}

\begin{figure}[t]
    \centering
    \footnotesize
    \subfigure[Corridor]{\includegraphics[width=0.24\linewidth]{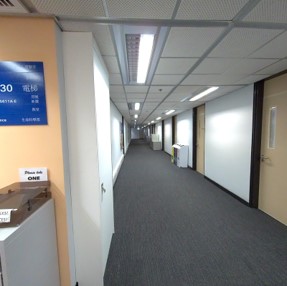}}
    \subfigure[Corridor]{\includegraphics[width=0.24\linewidth]{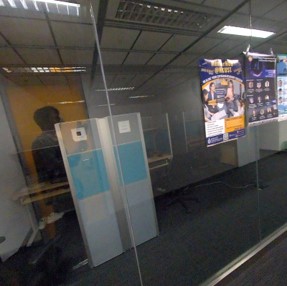}}
    \subfigure[Shared Office]{\includegraphics[width=0.24\linewidth]{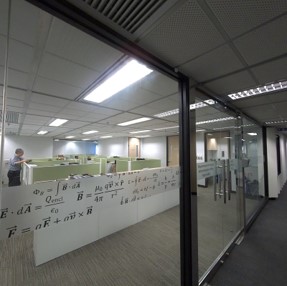}}
    \subfigure[Office]{\includegraphics[width=0.24\linewidth]{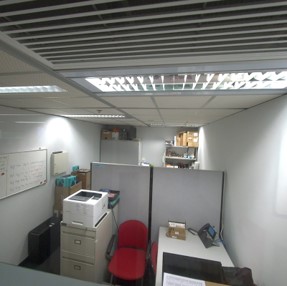}}
    \subfigure[Lift Lobby]{\includegraphics[width=0.24\linewidth]{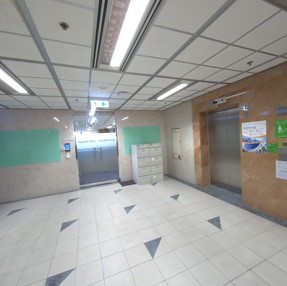}}
    \subfigure[Social Lobby]{\includegraphics[width=0.24\linewidth]{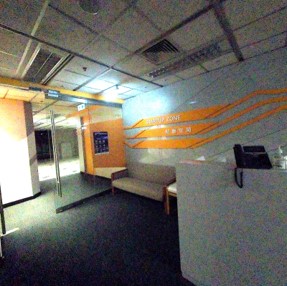}}
    \subfigure[Lounge]{\includegraphics[width=0.24\linewidth]{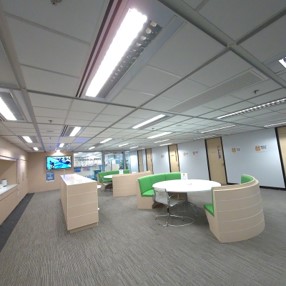}}
    \subfigure[Garden]{\includegraphics[width=0.24\linewidth]{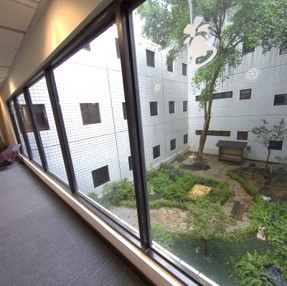}}
    \caption{Scenarios in the HKUST main (academic) building. Challenging conditions, such as long corridors and glass windows, pose difficulties for SLAM and applications with BIM.}
    \label{fig:scenes}
\end{figure}

\subsection{Multi-session Dataset and Data Structure}
\label{sec:multi-session}

\begin{figure}[t]
    \centering
    \includegraphics[width=\linewidth]{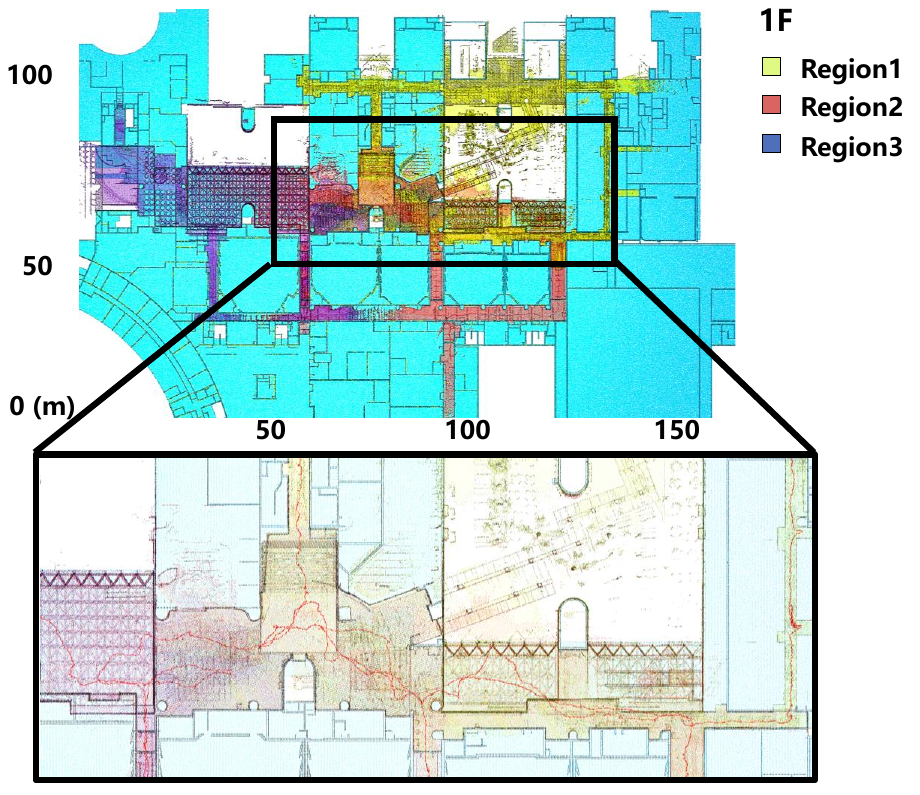}
    \caption{Trajectories and SLAM-generated maps of our self-collected session in the 1F of the HKUST BIM, shown with bird's eye view. The other sessions are distributed across different regions of the 3F, 4F and 5F. All trajectories can not be displayed due to the page limit, which will be open sourced for reference. }
    \label{fig:multi-session}
\end{figure}

We collect 12 sessions (sequences) across different floors and regions of the HKUST main building. These sessions encompass various indoor scenarios, as shown in Figure~\ref{fig:scenes}, presenting challenges for downstream applications. For instance, long and narrow corridors can cause LiDAR localization degeneration~\cite{tuna2023x}, while glass doors and walking pedestrians challenge visual perception and navigation. Additionally, deviations between as-built maps and as-designed BIM require high robustness for conventional algorithms. More detailed statistics for each session are presented in Table~\ref{tab:statistics}, in which Dis. is abbreviated for traveled distance, and Area is for covered area. Figure~\ref{fig:multi-session} shows the SLAM-generated LiDAR maps and trajectories on 1F, with a manual alignment on the floor plan exported from HKUST BIM.



\begin{table}[t]
\centering
\caption{Statistics of the Self-collected Muti-session Dataset}
\renewcommand\arraystretch{1.2}
\resizebox{\columnwidth}{!}{%
\begin{tabular}{cccccc}
\hline
\hline
Sequence & Time (s) & Dis. (m) & Area (m\textsuperscript{2}) & Num. of Scan & Num. of Image \\ 
\hline
1F-Region1 & 2238 & 422.79 & 2484 & 22471 & 537 \\
1F-Region2 & 1563 & 318.24 & 2053 & 15667 & 374 \\
1F-Region3 & 1001 & 235.76 & 1600 & 10051 & 240 \\ 
\hline
3F-Region1 & 2391 & 242.38 & 525  & 24006 & 576 \\
3F-Region2 & 1465 & 187.55 & 362  & 14704 & 347 \\
3F-Region3 & 1445 & 191.52 & 296  & 14503 & 347 \\ 
\hline
4F-Region1 & 1757 & 413.18 & 709  & 17640 & 419 \\
4F-Region2 & 1288 & 294.52 & 438  & 12935 & 312 \\
4F-Region3 & 501  & 134.12 & 271  & 5023  & 121 \\ 
\hline
5F-Region1 & 1401 & 246.25 & 604  & 14058 & 336 \\
5F-Region2 & 694  & 123.63 & 267  & 6965  & 164 \\
5F-Region3 & 680  & 162.84 & 401  & 6820  & 163 \\ 
\hline
\hline
\end{tabular}%
}
\label{tab:statistics}
\end{table}

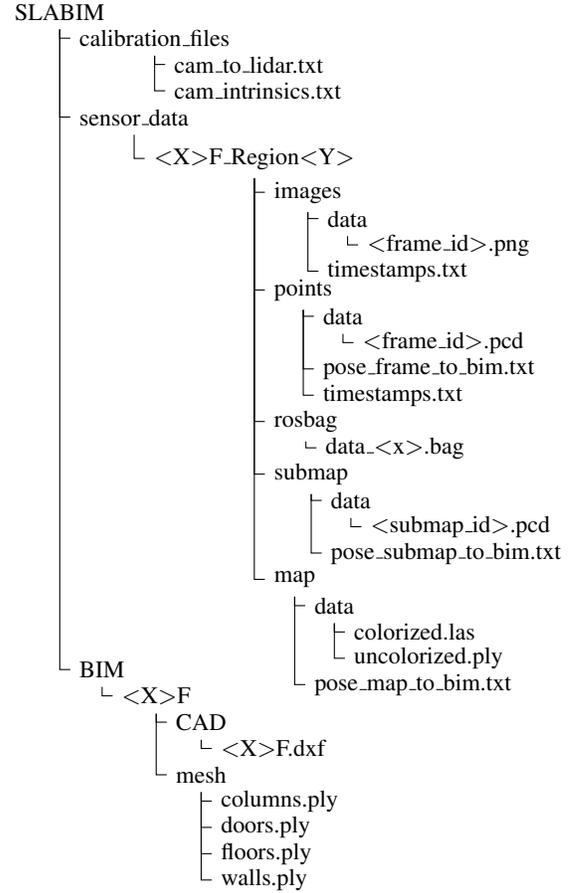
\begin{figure}[t]
    \centering
    \begin{tikzpicture}[
scale = 0.35,
every node/.style={font=\small, anchor=west, align=left},
  grow via three points={one child at (0.4,-1.0) and
  two children at (0.4,-1.0) and (0.4,-2.0)},
  edge from parent path={(\tikzparentnode.south) |- (\tikzchildnode.west)}]
  \node {SLABIM}
    child { node {calibration\_files}
      child { node {cam\_to\_lidar.txt} }
      child { node {cam\_intrinsics.txt} }
    }
    child [missing] {}
    child [missing] {}
    child { node {sensor\_data}
      child { node [yshift=-0.2cm] {$<$X$>$F\_Region$<$Y$>$}
        child { node [yshift=-0.1cm]{images}
          child { node {data}
            child { node {$<$frame\_id$>$.png} }
          }
          child [missing] {}
          child { node {timestamps.txt} }
        }
        child [missing] {}
        child [missing] {}
        child [missing] {}
        child { node {points}
          child { node {data}
            child { node {$<$frame\_id$>$.pcd} }
          }
          child [missing] {}
          child { node {pose\_frame\_to\_bim.txt} }
          child { node {timestamps.txt} }
        }
        child [missing] {}
        child [missing] {}
        child [missing] {}
        child [missing] {}
        child { node {rosbag}
          child { node {data\_$<$x$>$.bag} }
        }
        child [missing] {}
        child { node {submap}
          child { node {data}
            child { node {$<$submap\_id$>$.pcd} }
          }
          child [missing] {}
          child { node {pose\_submap\_to\_bim.txt} }
        }        
        child [missing] {}
        child [missing] {}
        child [missing] {}
        child { node {map}
          child { node {data}
            child { node {colorized.las} }
            child { node {uncolorized.ply} }
          }
          child [missing] {}
          child [missing] {}
          child { node {pose\_map\_to\_bim.txt} }
        }
      }
    }
    child [missing] {}
    child [missing] {}
    child [missing] {}
    child [missing] {}
    child [missing] {}
    child [missing] {}
    child [missing] {}
    child [missing] {}
    child [missing] {}
    child [missing] {}
    child [missing] {}
    child [missing] {}
    child [missing] {}
    child [missing] {}
    child [missing] {}
    child [missing] {}
    child [missing] {}
    child [missing] {}
    child [missing] {}
    child [missing] {}
    child { node {BIM}
      child { node {$<$X$>$F}
          child { node {CAD}
              child { node {$<$X$>$F.dxf} }
            }   
      child [missing] {}
      child { node {mesh}
        child { node {columns.ply} }
        child { node {doors.ply} }
        child { node {floors.ply} }
        child { node {walls.ply} }
        }
      }
    };
\end{tikzpicture}
    \caption{Data structure. SLABIM includes not only the HKUST BIM but also the self-collected multi-session data in the building. Illustrations of key directories are detailed in Section~\ref{sec:multi-session}.}
    \label{fig:datastructure}
\end{figure}

Figure~\ref{fig:datastructure} provides an overview of the data structure, including SLAM-oriented data and HKUST BIM. The following illustrations detail key functionalities:
\begin{itemize}

    \item \texttt{calibration\_files} provide intrinsic camera parameters and the extrinsic parameters to the LiDAR.
    
    \item In \texttt{sensor\_data/} directory, each session is named \texttt{<X>F\_Region<Y>}, with \texttt{X=1,3,4,5} and \texttt{Y=1,2,3} indicating the storey and region of collection, such as 3F\_Region1. This directory contains the \texttt{images} and \texttt{points} produced by camera and LiDAR; \texttt{data\_<x>.bag, x=0,1,2...} is the \texttt{rosbag} encoding the raw information, which can be parsed using ROS tools. 

    \item \texttt{sensor\_data/} also contains the maps generated by SLAM, including \texttt{submap} for the LiDAR-to-BIM registration (Section~\ref{sec:registration}) and optimized \texttt{map} by the offline mapping system. 

    \item \texttt{\small pose\_frame\_to\_bim.txt}, \texttt{\small 
 pose\_map\_to\_bim.txt} and \texttt{\small 
 pose\_submap\_to\_bim.txt} contains the ground truth poses from LiDAR scans and maps to the BIM coordinate. These poses are finely tuned using a manually provided initial guess and local point cloud alignment.
    
    
    \item \texttt{BIM/} contains CAD files (\texttt{.dxf}) and mesh files (\texttt{.ply}) exported from the original BIM models, organized by storey and semantic tags. Users can sample the meshes at specific densities to obtain point clouds, offering flexibility for various robotic tasks.
\end{itemize}

The SLABIM dataset is open-sourced with accompanying documentation\footnote{https://github.com/HKUST-Aerial-Robotics/SLABIM}. We recommend users customize the data from SLABIM based on practical needs. It is worth noting that FusionPortable\footnote{https://fusionportable.github.io/dataset/fusionportable/}~\cite{jiao2022fusionportable} and LOAM-Livox\footnote{https://github.com/hku-mars/loam\_livox}~\cite{lin2020loam} also partially contain sensor data collected in the HKUST main building. Intuitively, these external data can be utilized with the BIM model provided by the SLABIM in this paper. 



\begin{table}
\centering
\caption{Evaluation of Global LiDAR-to-BIM Registration on SLABIM}
\label{global-registration}
\renewcommand{\arraystretch}{1.2}
\resizebox{\columnwidth}{!}{%
\begin{tabular}{cc|cc|cc}
\hline
\hline
\multicolumn{2}{c|}{Method} & \multicolumn{2}{c|}{Pose Hough Transfrom \cite{qiao2024speak}} & \multicolumn{2}{c}{3D-BBS \cite{aoki20243d}} \\ \hline
\multicolumn{1}{c|}{Session} & Number & \multicolumn{1}{c|}{Recall (\%)} & Time (ms) & \multicolumn{1}{c|}{Recall (\%)} & Time (ms) \\ \hline
\multicolumn{1}{c|}{3F\_Region1} & 50 & 96.0 & 604.8 & 94.0 & 2806.5 \\
\multicolumn{1}{c|}{3F\_Region2} & 36 & 100.0 & 274.9 & 100.0 & 650.9 \\
\multicolumn{1}{c|}{3F\_Region3} & 42 & 95.2 & 188.9 & 100.0 & 2042.1 \\ \hline
\multicolumn{1}{c|}{4F\_Region1} & 111 & 87.4 & 298.6 & 96.4 & 1713.8 \\
\multicolumn{1}{c|}{4F\_Region2} & 69 & 97.1 & 208.0 & 91.3 & 3630.2 \\
\multicolumn{1}{c|}{4F\_Region3} & 27 & 100.0 & 517.1 & 100.0 & 2523.4 \\ \hline
\multicolumn{1}{c|}{5F\_Region1} & 62 & 98.4 & 393.3 & 100.0 & 581.3 \\
\multicolumn{1}{c|}{5F\_Region2} & 19 & 98.4 & 382.9 & 100.0 & 584.9 \\
\multicolumn{1}{c|}{5F\_Region3} & 39 & 87.2 & 276.2 & 100.0 & 586.0 \\ \hline
\multicolumn{1}{c|}{1F\_Region1} & 132 & 78.0 & 250.1 & 84.1 & 3920.6 \\
\multicolumn{1}{c|}{1F\_Region2} & 111 & 82.9 & 216.6 & 57.7 & 4301.4 \\
\multicolumn{1}{c|}{1F\_Region3} & 76 & 55.3 & 193.9 & 61.8 & 5879.6 \\ 
\hline
\hline
\end{tabular}
}
\label{tab:registration}
\end{table}

\setlength{\fboxsep}{0pt}
\begin{figure*}[t]
    \centering
    \footnotesize
    \subfigure[Success-1]{\fbox{\includegraphics[width=0.31\linewidth]{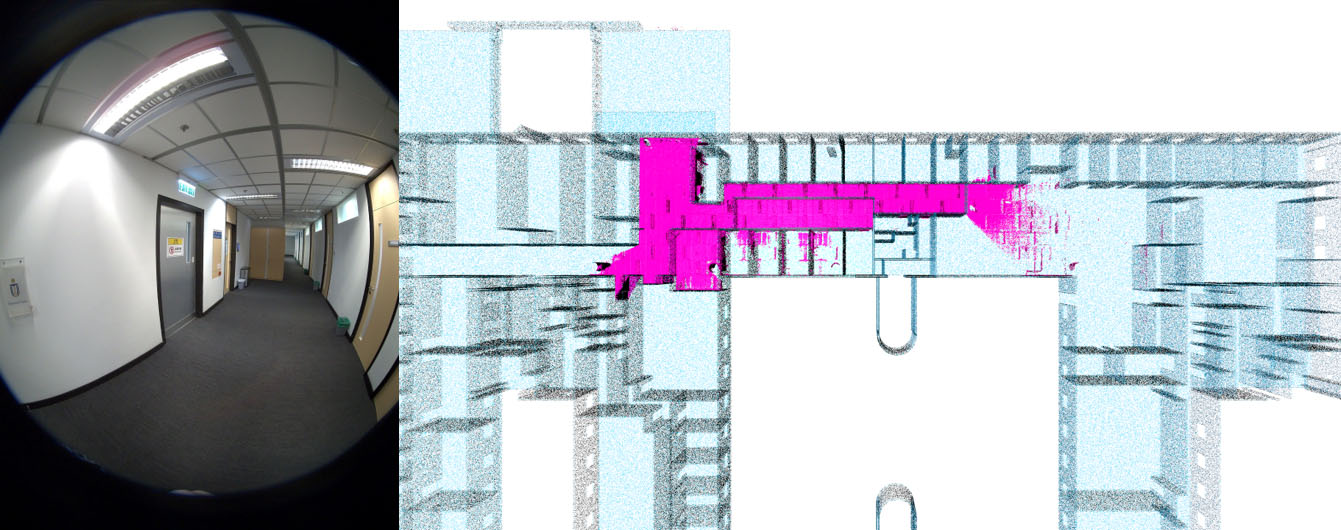}}}
    \subfigure[Success-2]{\fbox{\includegraphics[width=0.31\linewidth]{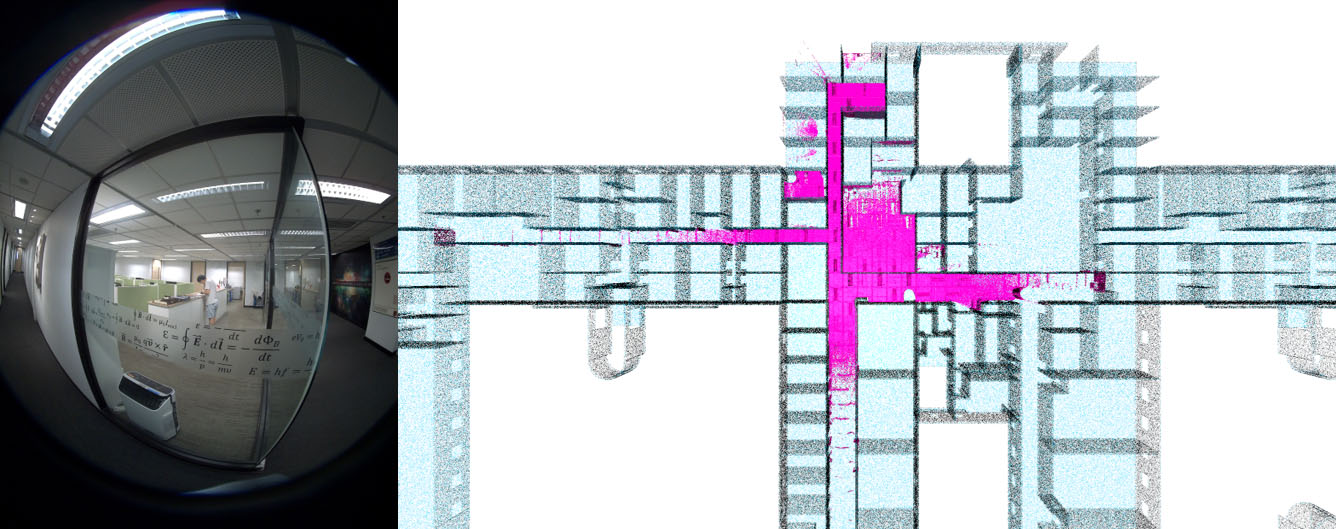}}}
    \subfigure[Success-3]{\fbox{\includegraphics[width=0.31\linewidth]{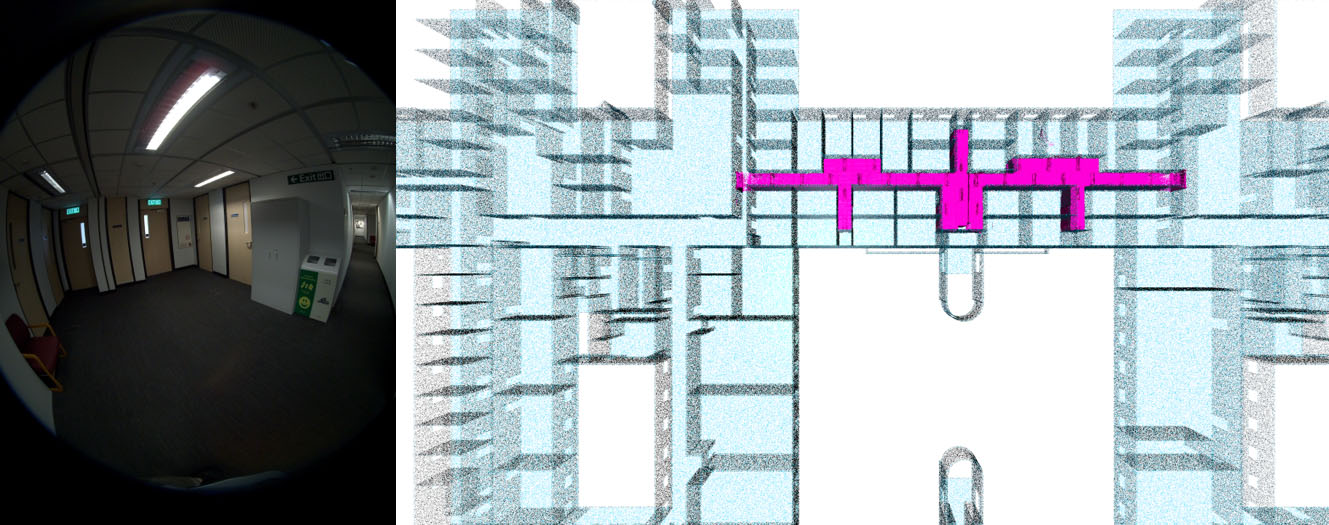}}}    
    \subfigure[Failure-1]{\fbox{\includegraphics[width=0.65\linewidth]{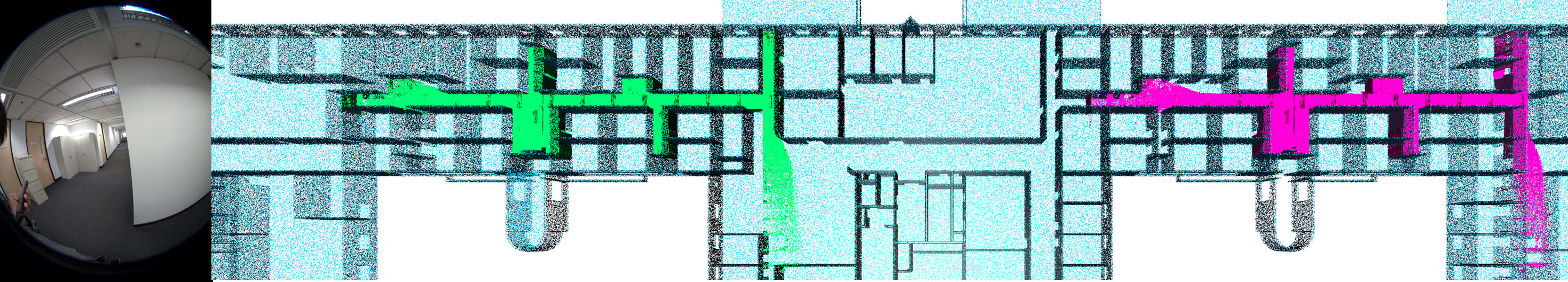}}}
    \subfigure[Failure-2]{\fbox{\includegraphics[width=0.295\linewidth]{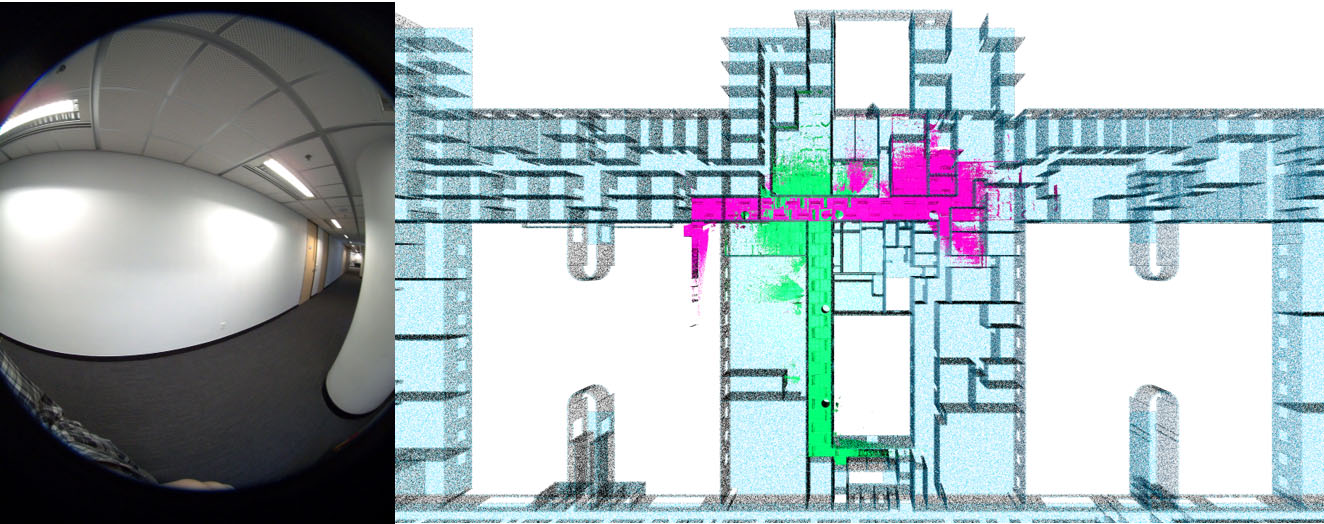}}} 

    \caption{Global LiDAR-to-BIM Registration results. Figures (a)-(c) are successful cases using the pose hough transform proposed in \cite{qiao2024speak}. Figures (g) and (h) show two failure cases where highly similar patterns exist in the building. Images from fish-eye cameras are also presented for better understanding.}
    \label{fig:registration}
\end{figure*}

\section{Experimental Validation and Evaluation} 
\label{sec:applications}

To demonstrate the practicality of SLABIM, we test three different tasks: 1) LiDAR-to-BIM registration, and 2) Robot pose tracking on BIM and 3) Semantic mapping evaluation. All tasks are achieved using advanced methods proposed in recent years.

\subsection{Global LiDAR-to-BIM Registration}
\label{sec:registration}

Global LiDAR-to-BIM registration aims to estimate a transformation from scratch between the LiDAR frame and the BIM coordinate system, which is critical for robotics and construction applications. For instance, an autonomous robot can localize itself globally by aligning LiDAR measurements to the BIM model without needing a pre-built map. Moreover, civil engineers can identify discrepancies between as-built and as-designed structures by comparing the point clouds to the BIM.

Despite its importance, global LiDAR-to-BIM registration presents several challenges as a cross-modal registration task, differing from traditional point cloud registration tasks. \textit{1) Modality Differences}: The inherent differences between LiDAR point clouds and BIM introduce complexities in feature extraction and matching. \textit{2) Repetitive Patterns}: Indoor environments often exhibit repetitive patterns, increasing association ambiguity. \textit{3) Building Deviations}: Discrepancies between as-built and as-designed structures create non-overlapping regions. These challenges also exist for visual images between BIM. Our dataset provides an experimental evaluation of global LiDAR-to-BIM registration. Considering the limited field of view indoors, we accumulate LiDAR scans into a submap with a path length of 30 meters, following the manner in our previous work~\cite{qiao2024speak}. The submaps and ground truth poses are also provided in the open-sourced SLABIM, as listed in Figure~\ref{fig:datastructure}.


In the SLABIM dataset, traditional handcrafted feature-based registration methods, such as FPFH-based matching combined with the TEASER estimator \cite{rusu2009fast,yang2020teaser}, have proven largely ineffective in these scenarios \cite{qiao2024speak}. Therefore, we select two recent advanced point cloud registration methods, 3D-BBS \cite{aoki20243d} and our previous study in \cite{qiao2024speak}, to test the LiDAR-to-BIM performance on SLABIM. Specifically, the 3D-BBS method employs a branch-and-bound (BnB) approach \cite{hess2016real} to search for the optimal transformation. The work by Qiao \textit{et al.} \cite{qiao2024speak} leverages common structural elements of buildings to construct a triangular descriptor and estimate the transformation using pose hough transform. The results are summarized in Table~\ref{tab:registration}. Both methods achieve satisfactory performance, though there remains room for improvement regarding recall and speed. Figure~\ref{fig:registration} presents case studies using our work \cite{qiao2024speak}. Moreover, future directions include deep learning-based approaches, semantic-enhanced registration, and 2D image-to-3D BIM alignment.



\subsection{Robot Pose Tracking on BIM}

The experiments above focus on registering 3D point clouds to BIM globally without an initial guess. For mobile robots, continuous pose tracking is essential to support autonomous navigation. Pose tracking on BIM encounters similar challenges as listed in Section~\ref{sec:registration}. However, the problem formulation differs: LiDAR-to-BIM aims to estimate the one-shot robot pose for global localization, while pose tracking requires estimating poses given the initial state and sequential measurements. The problem formulation is detailed in a recent survey paper~\cite{yin2024survey}.

To achieve pose tracking, the HKUST BIM is converted into a global point cloud map sampled from the mesh. We design a simple yet effective optimization-based method to track the robot (LiDAR sensor). Specifically, the system maintains a local point cloud map using a sliding window, which contains several keyframes and computes odometry using generalized iterative closest point (GICP)~\cite{koide2021voxelized}. The estimated pose from odometry can then be corrected using a LiDAR-to-BIM measurement model. Considering inconsistencies between the BIM and the actual environment, our optimization formulation includes a prior pose factor and several LiDAR-to-BIM factors using point-to-plane distance with relatively strict bounds. Although the method heavily relies on LiDAR measurements and alignment, which may lead to overconfidence for state estimation, it effectively supports pose tracking on BIM-generated maps.



\begin{figure}[t]
    \centering
    \includegraphics[width=0.99\linewidth]{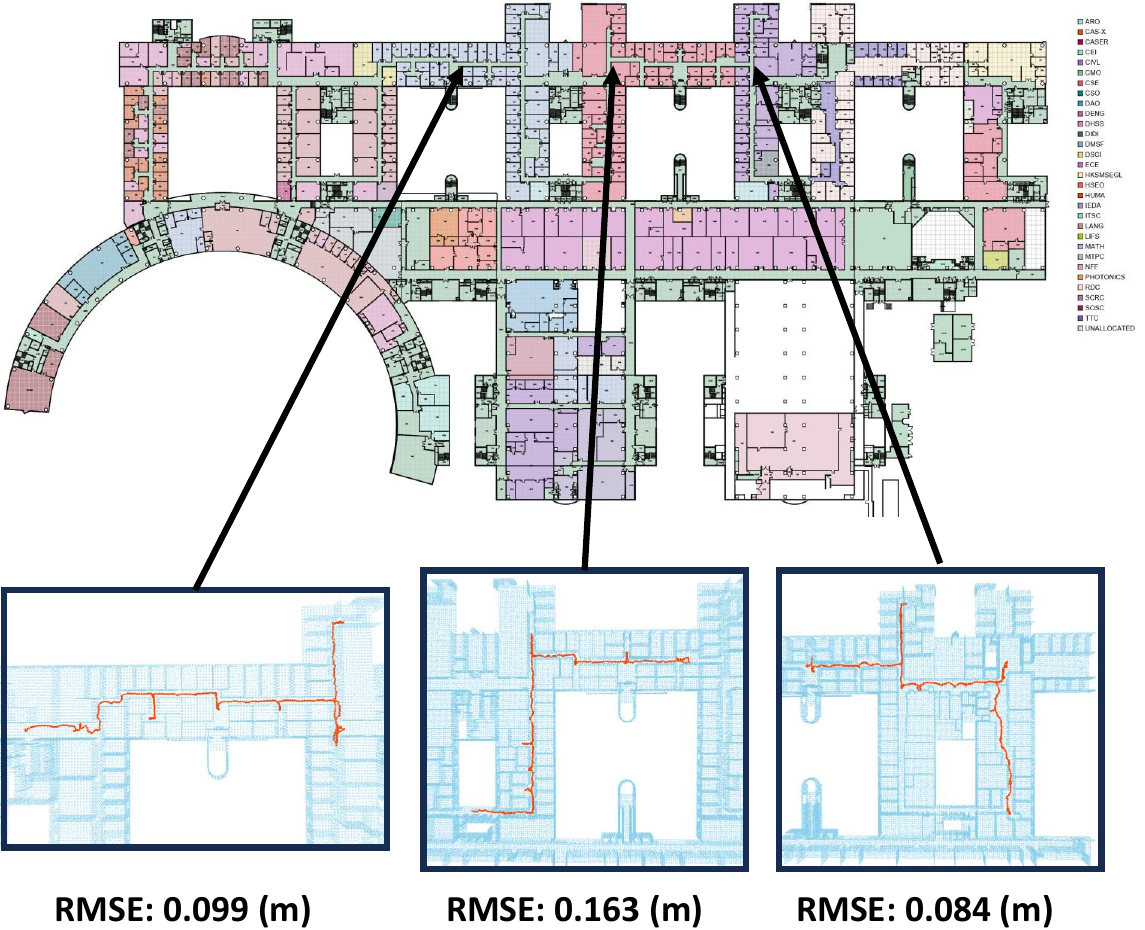}
    \caption{Trajectories of pose tracking (in red) on BIM-generated 3D point cloud map (in blue), including place annotations on the 3F floor plan. We first convert BIM to point cloud maps, then use sequential LiDAR measurements as input. Though deviations exist between BIM and as-built world, the designed Kalman filter successfully tracks the pose, demonstrating the feasibility of using BIM as a map.}    
    \label{fig:pose_tracking}
\end{figure}

Figure~\ref{fig:pose_tracking} shows the tracked robot poses. We also evaluate the errors compared to the ground truth poses provided by the offline mapping system provided by the sensor suite. The quantitative results demonstrate that BIM-generated maps could be a viable choice for robot localization. Visual localization on building architecture is also a promising direction for more generalized applications~\cite{chen2024f3loc}, such as indoor augmented reality (AR).

\subsection{Semantic Mapping Evaluation}


Semantic mapping is a popular topic in mobile robotics. Advanced methods, e.g., Kimera~\cite{rosinol2020kimera} and our recent FM-Fusion~\cite{liu2024fm}, can generate a metric-semantic map or high-level scene graph to model the environments. Evaluating semantic mapping in indoor environments requires a ground-truth semantic map. However, obtaining ground-truth semantic annotation at building scale is challenging due to the labor-intensive nature of manual annotation. Conversely, BIM provides semantic labels that can be used for ground truth generation, thus making semantic mapping evaluation more feasible.

\begin{figure}[t]
    \centering
    \footnotesize


    \subfigure[Scenario-1]{\includegraphics[width=0.95\linewidth]{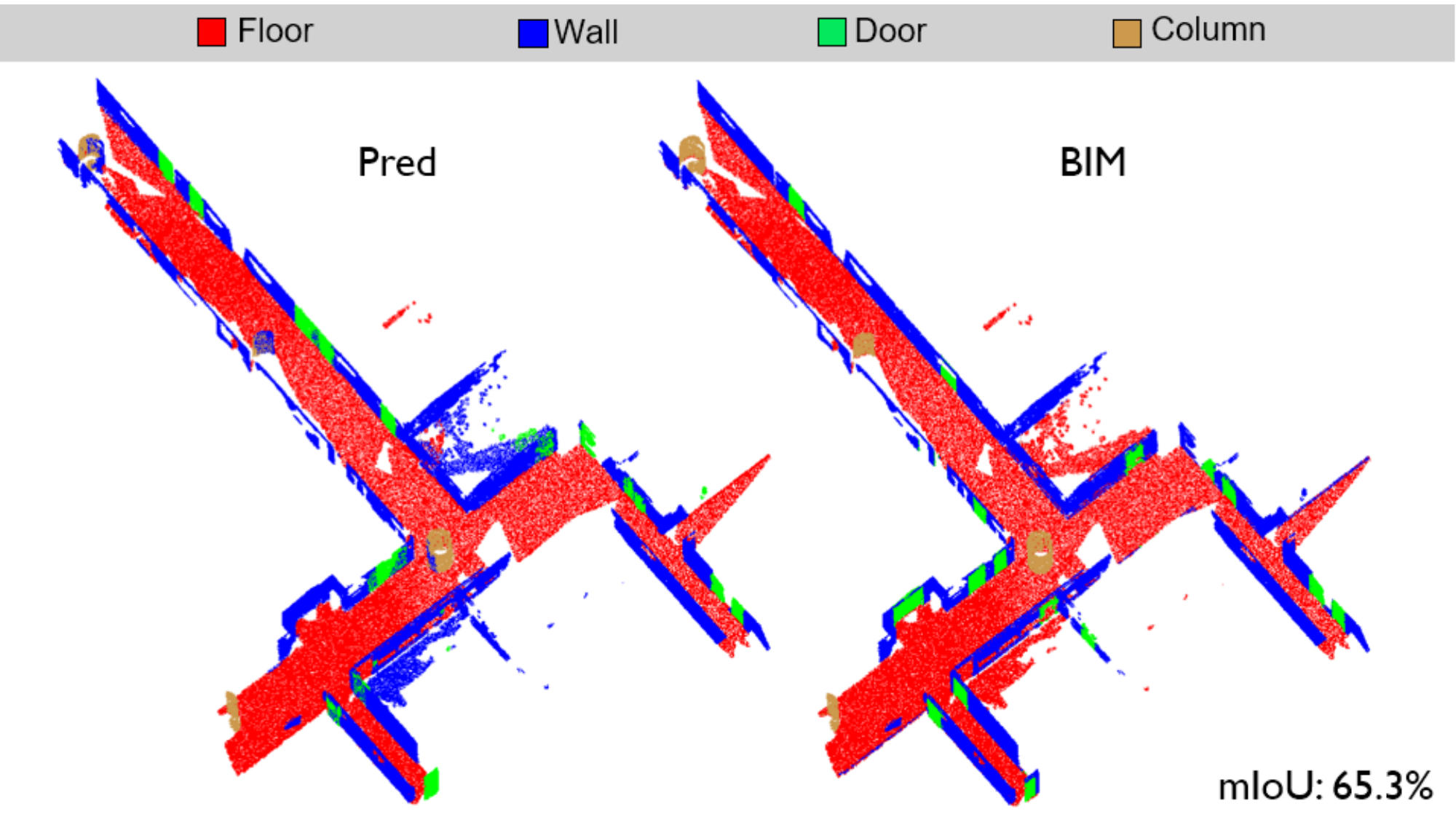}}
    \subfigure[Scenario-2]{\includegraphics[width=0.95\linewidth]{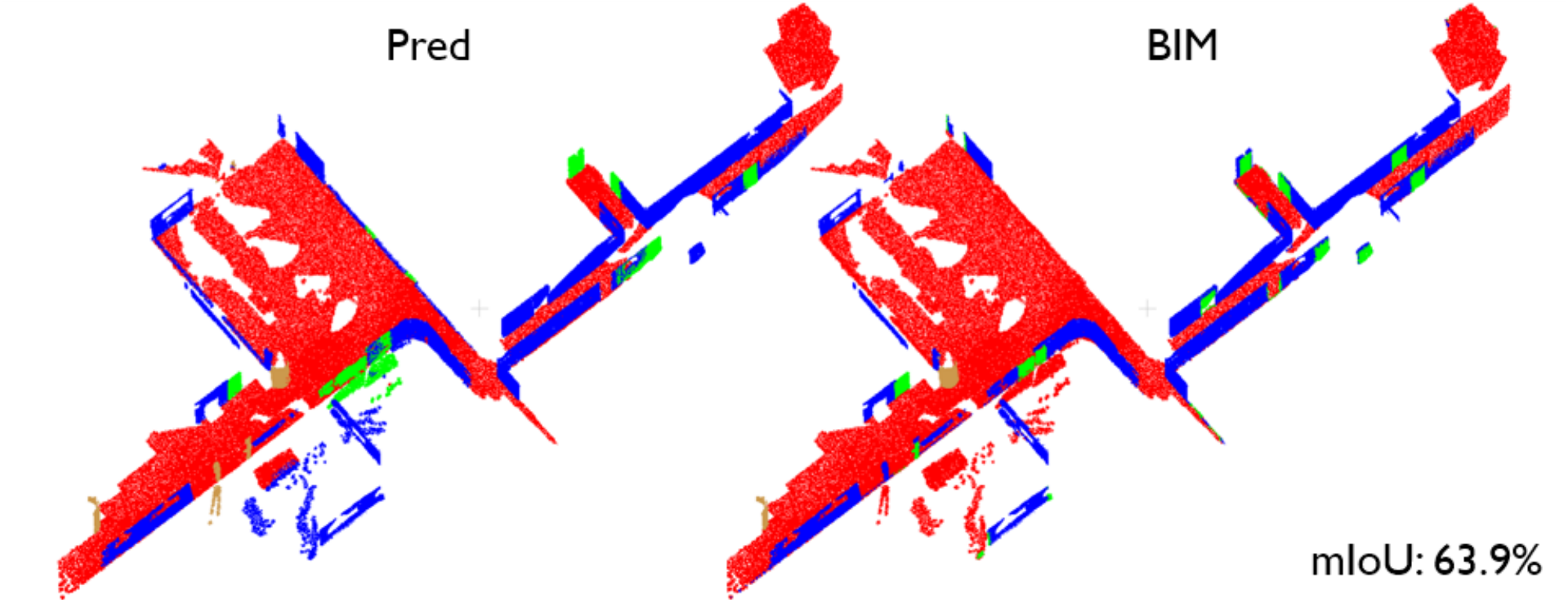}}

    \caption{Visualization of semantic maps at two scenarios from SLABIM. The predicted (Pred) semantic map from the FM-Fusion~\cite{liu2024fm} is shown on the left. The ground-truth annotation from BIM is shown on the right.}
    \label{fig:semantic_mapping}
\end{figure}


To achieve this, we deploy our FM-Fusion \cite{liu2024fm} using the self-collected data in SLABIM. FM-Fusion integrates visual segmentation results from foundation models \cite{kirillov2023sam} for semantic mapping. For the ground truth, we convert the HKUST BIM into semantic point cloud maps using labeled information. Specifically, we project the annotated labels from BIM onto the predicted point cloud maps from FM-Fusion. This projection generates a ground-truth semantic map in the same Cartesian coordinate as our predicted one. Both maps contain four semantics: floor, wall, door, and column, the common elements in indoor environments.


As shown in Figure~\ref{fig:semantic_mapping}, the predicted semantic map from FM-Fusion is mostly consistent with the ground-truth map annotated using BIM. For quantitative evaluation, we follow the intersection-over-union (IoU) metric in ScanNet~\cite{dai2017scannet}. The IoU metrics of all sequences are summarized in Table~\ref{tab:iou}. FM-Fusion achieves a high IoU in two semantic types: floor and wall. Due to the frequent partial view of images, GroundingDINO \cite{liu2023grounding} detects doors and columns with lower accuracy, resulting in a lower IoU for these elements. One might argue that using BIM as the ground truth is unreasonable due to environmental changes, which can cause deviations between as-built and as-designed states. In this paper, we assume these changes are sufficiently minor and focus primarily on the dataset validation.


\begin{table}[t]
    \centering
    \caption{Evaluation of Semantic Mapping by IoU}
    \begin{tabular}{c c c c c c}
        \hline
        \hline
         & Floor & Wall & Door & Column & mIoU \\
        \hline
        IoU & 85.2 & 71.9 & 44.3 & 31.6 & 58.3 \\
        \hline
        \hline
    \end{tabular}
    \label{tab:iou}
\end{table}

\section{Conclusions}

We present SLABIM, the \textit{first} SLAM-BIM coupled dataset. SLABIM is constructed from a university BIM model and incorporates multi-session multi-sensor data. We decompose the BIM into other formats and modalities for ease of use. Our self-collected data is analyzed and organized for users. Furthermore, this paper reports experimental validation and evaluation using SLABIM. Future work will focus on dataset expansion and iteration, along with in-depth studies on robotics-enhanced construction automation. We hope SLABIM will promote interdisciplinary research within the community.

\section*{Acknowledgement}
We sincerely thank Prof. Jack C. P. Cheng for providing the original HKUST BIM files. 


\bibliographystyle{IEEEtran}
\bibliography{root}

\end{document}